%% file: main.tex
\documentclass[10pt,twocolumn,letterpaper]{article}
\pdfoutput=1

\usepackage[pagenumbers]{wacv} 

\usepackage{graphicx}
\usepackage{amsmath}
\usepackage{amssymb}
\usepackage{booktabs}
\usepackage{algorithm}
\usepackage{algorithmic}
\usepackage{caption}
\usepackage{subcaption}
\usepackage{makecell} 
\newcommand{\cmark}{\ding{51}}
\newcommand{\xmark}{\ding{55}}
\usepackage{times}
\usepackage{multirow,multicol}
\usepackage{epsfig}
\usepackage{graphicx}
\usepackage{multirow}
\usepackage{pifont}
\usepackage{paralist}

\newcommand{\walon}[1]{\textcolor{black}{#1}}

\usepackage{xcolor}
\usepackage{colortbl}
\definecolor{col1}{RGB}{232, 161, 148}
\definecolor{col2}{RGB}{148, 187, 232}

\usepackage[pagebackref,breaklinks,colorlinks]{hyperref}

\usepackage[capitalize]{cleveref}
\crefname{section}{Sec.}{Secs.}
\Crefname{section}{Section}{Sections}
\Crefname{table}{Table}{Tables}
\crefname{table}{Tab.}{Tabs.}

\def\wacvPaperID{2227} 
\def\confName{WACV}
\def\confYear{2024}

\begin{document}

\title{Masking Improves Contrastive Self-Supervised Learning for ConvNets, \\ and Saliency Tells You Where\\}

\author{Zhi-Yi Chin$^{1}$\protect\footnotemark[1] \ \ Chieh-Ming Jiang$^{1}$\protect\footnotemark[1] \ \ Ching-Chun Huang$^{1}$ \ \ Pin-Yu Chen$^{2}$ \ \ Wei-Chen Chiu$^{1}$  \\
        {}$^1$ National Yang Ming Chiao Tung University \\
        {}$^2$IBM Research\\
        {\tt\small \{joycenerd.cs09, nax1016.cs10, chingchun\}@nycu.edu.tw, pin-yu.chen@ibm.com, walon@cs.nctu.edu.tw}\\
}

\maketitle
\renewcommand{\thefootnote}{\fnsymbol{footnote}}
\footnotetext[1]{These authors contributed equally to this work}

\begin{abstract}
   \walon{While image data starts to enjoy the simple-but-effective self-supervised learning scheme built upon masking and self-reconstruction objective thanks to the introduction of tokenization procedure and vision transformer backbone, convolutional neural networks as another important and widely-adopted architecture for image data, though having  contrastive-learning techniques to drive the self-supervised learning, still face the difficulty of leveraging such straightforward and general masking operation to benefit their learning process significantly. In this work, we aim to alleviate the burden of including masking operation into the contrastive-learning framework for convolutional neural networks as an extra augmentation method. In addition to the additive but unwanted edges (between masked and unmasked regions) as well as other adverse effects caused by the masking operations for ConvNets, which have been discussed by prior works, we particularly identify the potential problem where for one view in a contrastive sample-pair the randomly-sampled masking regions could be overly concentrated on important/salient objects thus resulting in misleading contrastiveness to the other view. To this end, we propose to explicitly take the saliency constraint into consideration in which the masked regions are more evenly distributed among the foreground and background for realizing the masking-based augmentation. Moreover, we introduce hard negative samples by masking larger regions of salient patches in an input image. Extensive experiments conducted on various datasets, contrastive learning mechanisms, and downstream tasks well verify the efficacy as well as the superior performance of our proposed method with respect to several state-of-the-art baselines.}
\end{abstract}

\input{intro.tex}
\input{related.tex}

\input{method.tex}
\input{exp.tex}

\input{conclusion.tex}

\clearpage
\input{supp.tex}

{\small
\bibliographystyle{ieee_fullname}
\bibliography{egbib}
}

\end{document}

%% file: intro.tex
\section{Introduction}\label{sec:intro}
\vskip -0.1cm
\walon{
The recent renaissance of deep learning techniques has brought a magic leap to various fields, such as computer vision, natural language processing, and robotics. Learning from a large-scale labeled/supervised dataset, which is one of the key factors leading to the success of deep learning, however, has now turned out to be a significant limitation on its extensions to more fields. In addition to the expensive cost of time and human resources to collect training datasets for different tasks and their corresponding labels, the supervised learning scenario typically would suffer from the issue of overfitting on the training dataset, thus leading to worse generalizability of the learnt models. These problems bring challenges for the application of deep learning techniques but also give rise to the research topic of self-supervised learning, wherein it aims to learn to extract informative feature representations from an unlabelled dataset via leveraging the underlying structure of data and building the supervisory signals from the data itself. The discovered representations are typically more general and can be further utilized or fine-tuned to various downstream tasks.}

\walon{Without loss of generality, self-supervised learning has firstly made great success in natural language processing, where the autoregressive modeling (i.e. predicting the next word given the previous words) and masked modeling (i.e. masking operation to randomly mask a portion of words in a text, coupled with a self-reconstruction objective to predict those masked words) bring up the powerful language models such as GPT~\cite{radford2018improving} and BERT~\cite{devlin2018bert}. Nevertheless, the direct adaptation of such techniques (especially masking and self-reconstruction) to image data~\cite{pathak2016context} only contributes to slight improvement (at least not as significant as what happens in the field of natural language processing), in which such a predicament was later relieved with the help of vision transformers~\cite{dosovitskiy2020image} (e.g. the influential work from masked autoencoder (MAE)~\cite{he2022masked} and the related ones such as SimMIM~\cite{xie2022simmim}, BEiT~\cite{bao2021beit}, and ~iBOT \cite{zhou2021ibot}). In contrast to vision transformers which enable the application of masking operation and its coupled self-reconstruction objective on the self-supervised learning for vision data, another dominant architecture over the last decade for computer vision field, i.e. convolutional neural networks, has difficulty incorporating the random masking operation (on image patches), because the resultant edges between masksed and unmasked regions could cause problems for learning convolution kernels, and the nature of performing convolutions on regular grids also hinder it from adopting positional embeddings or masked tokens as the typical transformer models~\cite{he2022masked}.}

\walon{In turn, the most popular self-supervised learning scenarios nowadays for convolution neural networks come from contrastive learning -- given one sample image, two different views of it are respectively created by two different augmentations. The contrastive objective which attracts the views from the same image (known as positive pair/views) while repelling the ones from distinct images (respectively, negative pair/views) drives the learning of feature extractor (i.e. encoder) to capture the crucial features invariant to augmentations. Hence, how to design good positive and negative views with augmentations~\cite{peng2022crafting, ge2021robust} plays an important role in the success of contrastive learning, in which the design choices for augmentations typically are highly dependent on the characteristics of the image data domain (i.e. more domain-specific). From such a point of view, including the less domain-specific augmentations would definitely be able to benefit the versatility and the flexibility of the corresponding contrastive learning algorithms, thus the fundamentals of  masking (as being one of the most straightforward and general operations) consequently come into our sight: \textit{Are we able to include masking as an extra augmentation method into contrastive self-supervised learning framework with convolutional neural networks as its backbone?}
}

\walon{We are not the first to ask such a question. Two prior works (i.e. \textbf{MSCN}~\cite{jing2022masked} and \textbf{ADIOS}~\cite{shi2022adversarial}) proposed to tackle the issues ``how to mask'' and ``learning where to mask'' respectively. Nevertheless, on one hand, the improvements provided by these prior works are relatively insignificant, thus showing this topic is still under-explored; on the other hand, we highlight the potential issue that: if the masking is performed in a completely random manner, there exists a chance where all the masked patches fall on either the foreground or the background objects, in which the contrastive objective upon positive pair under such case could be detrimental for the overall model learning (e.g. attracting two views where one is completely background while the other still owns most of the foreground).}

\walon{To address this potential issue, we propose to particularly include \textbf{saliency} as a prior before performing masking. That is, we suggest that the masked patches should be evenly distributed to an image's foreground objects and background, regardless of the masking ratio. To this end, we introduce \textbf{\textit{random masking with saliency constraint}} as an augmentation method for the contrastive self-supervised learning framework, in which the feature extractor is built upon convolutional neural networks. Basically, we split the entire input image into the foreground objects and the background, followed by performing random masking on them separately, and three different masking strategies are provided to handle the parasitic edges stemming from masking. Moreover, we also introduce hard negative samples by masking more salient patches of the original input image, where these hard negative samples are experimentally shown to bring an extra boost to our proposed method. Lastly, we further discover that masking only one branch (to be specific, masking only the query branch when processing the positive pairs) of the contrastive learning framework (usually also known as \textit{siamese network}) provides better performance than masking both branches due to the effects in terms of sample variance that masking brings, which also well corroborates the statement claimed in~\cite{wang2022importance}. Our main contributions in this work are summarized as follows,
\begin{compactitem}
\item We propose a saliency masking augmentation method for the contrastive self-supervised learning framework with convolutional neural networks as backbones, where the saliency information is utilized to guide the random masking applied on the foreground and background regions individually.
\item Three masking strategies are proposed to tackle parasitic edges between masked and unmasked regions, in which hard negative samples can also be created by masking more salient patches to achieve further improvement.
\item From the perspective of manipulating the difference in terms of variance between two branches of the siamese network, we propose to apply masking augmentation solely on the query branch when processing positive pairs to benefit the model training.
\end{compactitem}
}

%% file: related.tex
\section{Related work}\label{sec:related}
\vskip -0.1cm
\noindent\walon{\textbf{Self-Supervised Learning} (SSL) aims to learn a feature encoder for extracting representations from unlabeled data via the help of pretext tasks (where the objective functions for these tasks are typically built upon the data properties), in which the resultant encoder can be further fine-tuned with labeled data to support different downstream tasks. Early SSL works rely on designing handcrafted pretext tasks, such as predicting rotation angles~\cite{gidaris2018unsupervised, feng2019self}, solving jigsaw puzzles~\cite{noroozi2016unsupervised}, or colorization~\cite{zhang2016colorful}. Recently, the introduction of contrastive objectives to SSL algorithms~\cite{chen2020simple, he2020momentum, grill2020bootstrap, caron2020unsupervised, zbontar2021barlow} have brought a significant leap of performance, even providing superiority to some standard supervised learning baselines. Contrastive SSL tries to maximize the agreement between representations of positive samples (i.e. augmented views from the same source image). While some contrastive SSL approaches (such as SimCLR \cite{chen2020simple} and MoCov2 \cite{he2020momentum}) further leverage negative samples (i.e., augmented views from different images) to prevent the model collapse (i.e. learning trivial features) by utilizing large batch size and memory bank, some other works (e.g. SimSiam~\cite{chen2021exploring} and BYOL~\cite{grill2020bootstrap}) instead prove that the negative samples might not be necessary for contrastive SSL (e.g. the stop-gradient technique serve the same purpose to prevent model collapse). Though there exist other categories of SSL methods (e.g. clustering ones such as DeepCluster~\cite{caron2018deep} and SWAV~\cite{caron2020unsupervised}), the contrastive ones still take the lead in the stream of SSL.}

\noindent\walon{\textbf{Masking in SSL} (e.g. masking out a portion of input data sample followed by learning to recover the missing content) is firstly proved by the success of masked language modeling (e.g. BERT~\cite{devlin2018bert}) and later adapted to the vision data, thanks to the introduction of vision transformer backbones where the input images are firstly divided into patches then tokenized. For instance, MAE~\cite{he2022masked} as a seminal work proposes an autoencoder architecture where the transformer-based encoder turns the unmasked image patches into feature representations, which are further decoded back to the original image; while SimMIM \cite{xie2022simmim} encodes the entire image, including the masked patches, and predicts the missing region with a lightweight one-layer head. Compared to the SSL methods for vision data which are based on the coupled masking operation and self-reconstruction loss but highly constrained to the transformer backbone, contrastive SSL becomes more friendly for adopting another important computer vision backbone, convolutional neural networks (also abbreviated as ConvNets), in which recently there comes some research works to investigate the plausibility of including masking operation as an augmentation method into contrastive SSL for ConvNets (also denoted as ``siamese networks with ConvNets'' in this paper).  
}

\walon{MSCN~\cite{jing2022masked} firstly discusses the issues of adopting masking operation in siamese networks with ConvNets (including the parasitic edges on the masked input, introduction of superficial solutions, distortion upon the balance between local and global features, and having fewer training signals) then proposes several designs to tackles these issues, such as adopting high-pass filter to alleviate the impact of parasitic edges or applying focal masks to balance short-range and long-range features. Basically, MSCN focuses more on the perspective of ``how to mask''. In comparison, our work not only provides more ``how to mask'' strategies (in addition to the one using a high-pass filter, two more based on strong blurring and filling mean value are proposed) but also explicitly includes saliency constraint (from the perspective of ``where to mask'') as well as extensions on learning mechanism (i.e. using masking to produce hard negative samples and manipulate variance across siamese branches); ADIOS~\cite{shi2022adversarial} mainly tackles the issue of ``where to mask'', where instead of using random masking, it particularly adopts an occlusion module (UNet-based, acting as a segmentation model) which learns adversarially along with the feature encoder to determine the regions to be masked, hence the produced masks are semantically meaningful. As jointly training the feature encoder and occlusion module results in heavy computational cost for ADIOS, our proposed method strikes a better balance between having (partially) semantic masks (as being guided by saliency to separate foreground and background) and the computation efforts (as our saliency is estimated by a pretrained and frozen localization network). The visualization to highlight the difference among our proposed method, MSCN~\cite{jing2022masked}, and ADIOS~\cite{shi2022adversarial} is provided in Figure~\ref{fig:model_comparison}. 
}

%% file: method.tex
\begin{figure*}[ht]
\centering
\includegraphics[width=1\textwidth]{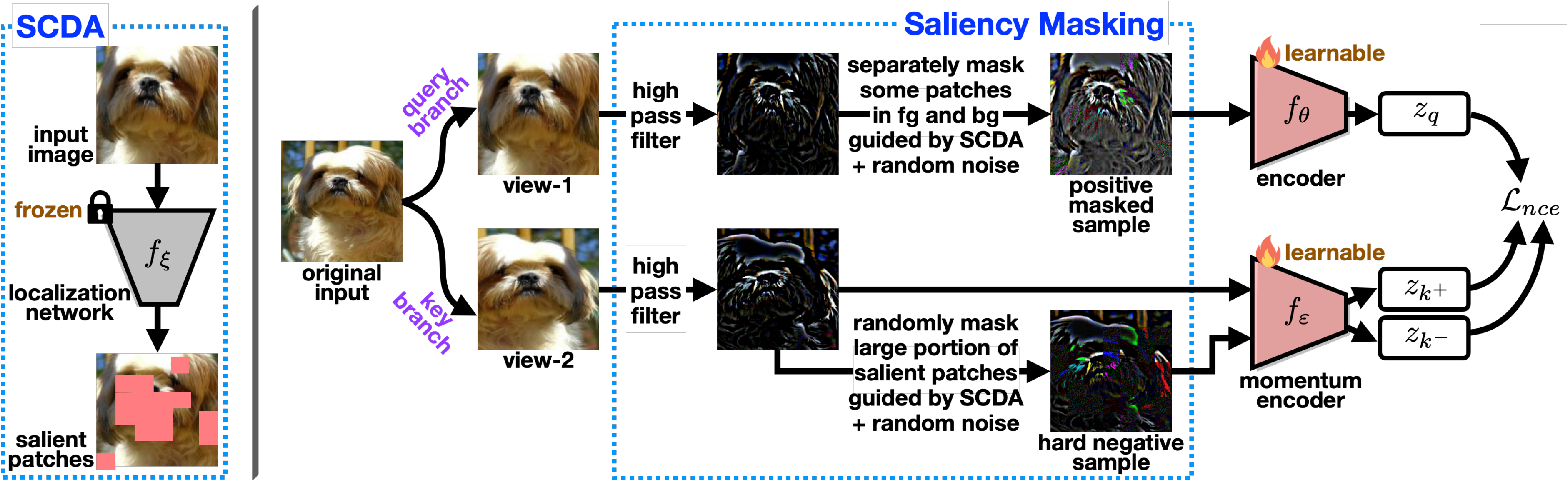}
\vspace{-1em}
\caption{\walon{An overview for our proposed method of including saliency-guided masking augmentation into contrastive self-supervised learning, where the backbone of the feature extractor is ConvNets. Firstly, our localization network $f_{\xi}$ produce the saleincy map which is built upon SCDA~\cite{wei2017selective} (cf. Section~\ref{Object localization}), in which such saliency map helps to separate the foreground objects and background in an image. Given an input image, after conducting standard augmentations along the query and key branches (following the common practice of siamese network) to produce two views, our proposed saliency-guided masking strategies are adopted to produce positive and hard negative samples (please refer to Section~\ref{Masking strategy} for more details, where in this figure we take the high-pass filtering strategy as an example). The constructed positive and (hard) negative samples are gone through the feature encoder to compute the contrastive objective function $\mathcal{L}_{nce}$ (please refer to our Section~\ref{Overall design}). Noting that here we base on the SSL framework of MoCov2~\cite{chen2020improved} to illustrate the computation flow, hence there exists an momentum encoder $f_{\varepsilon}$ in addition to our main learning target, the feature encoder $f_{\theta}$. }
}
\vspace{-1.5 em}
\label{framework}
\end{figure*}

\section{Method}\label{sec:method}
\vskip -0.1cm

\walon{As motivated previously, we would like to include the masking operation as an extra augmentation method into contrastive self-supervised learning with ConvNets as backbone, where the saliency information is particularly leveraged to guide the masking. Our full model is shown in Figure~\ref{framework} where in the following we will sequentially describe the saliency computation, our various saliency-guided masking strategies, the ways to construct positive and hard negative samples, as well as the learning scheme. 
}

\subsection{Saliency computation}\label{Object localization}
\vskip -0.1cm
\walon{The idea of saliency was firstly introduced to predict the eye-catching regions over an image, in which here we generalize such idea to localize the main objects in an input image (which are corresponding to ``foreground'' without loss of generality) while the rest is then treated as ``background''. To this end, in this work we adopt the Selective Convolutional Descriptor Aggregation (\textbf{SCDA}~\cite{wei2017selective}) method to build our \textit{localization network} $f_{\xi}$ for producing saliency map $M$ of a given image $X$, i.e. $M = f_{\xi}(X)$. The main reason to choose SCDA for our use stems from its simplicity of only requiring a pre-trained CNN model (typically for the task of classification) and demanding no further supervision to localize the main objects. With denoting the feature tensor obtained by the aforementioned pre-trained CNN model prior to its global average pooling layer as $S \in \mathbb{R}^{U\times V\times D}$, the aggregation (by summation) of $S$ along the channel dimension results in the activation map $A \in \mathbb{R}^{U\times V}$. In addition to use the mean value $\Bar{a}$ of $A$ as the threshold on all the $U\times V$ elements in $A$ to locate the positions of foreground objects (same as the original SCDA), we add another condition based on the standard deviation $\sigma$ of $A$ to have more flexible localization results which better fit our need to later guide the masking:
\begin{equation}
\label{eq:saliency}
    M(u,v)=
        \left\{
            \begin{aligned}
                1 &\quad\text{if } A(u,v) \geq \Bar{a} - 0.6\cdot\sigma 
                \\
                0 &\quad\text{otherwise} 
            \end{aligned}
        \right.
\end{equation}
Take an input image of size $224\times 224$ and use an ImageNet-pretrained CNN model based on ResNet-50 as an example, the resultant saliency map $M$ is of size $7\times 7$ and every its element is corresponding to a $32\times 32$ patch in the original image, in which such image patches are the basic units for our performing masking operation later in constrative SSL.
}

\subsection{Saliency-guided masking strategies}\label{Masking strategy}
\vskip -0.1cm
\walon{As naively masking out image patches in an original image will produce many parasitic edges (i.e. the boundaries between masked and unmasked patches/regions), the ConvNet-based feature encoder could be largely misled to focus on learning these unwanted edge features (since the convolutional kernels are typically good at capturing edges) thus causing problematic model training. In this work we propose to adopt three masking strategies for amending the aforementioned issue stemmed from parasitic edges:
\begin{compactitem}
  \item \textbf{High-pass filtering.} Such masking strategy is actually proposed by MSCN~\cite{jing2022masked},  where the high-pass filter is applied on the input image prior to the masking operation, in which the edges caused by masking in the filtered map is less visible. It is worth noting that, as the input for the feature encoder under such masking strategy is the high-pass-filtered images, in the downstream tasks the input data should follow the same form (i.e. needed to be firstly high-pass-filtered as well) to achieve better performance, in which this requirement would become a limitation for practical applications (i.e. users for the downstream tasks have to know how the encoder was trained in SSL stage).
  \item \textbf{Strong blurring.} The masking is performed firstly on the original input image, where the masked regions are not filled with the zero value but their own appearance processed by strong Gaussian blurring (i.e. each region to be masked is gone through a low-pass filter), leading to less obvious parasitic edges. Noting that within the patches/regions undergone such masking strategy, the image details are also lost while only the significant contours of objects are preserved. The Gaussian blurring kernel is of size $31\times 31$ with variance set to $10$ in our experiments unless otherwise stated. 
  \item \textbf{Mean filling.} The masking is also performed on the original input image at first, then the masked regions are filled by the mean pixel value of that input image, such that the boundaries between the masked and unmasked regions becomes much less significant.
\end{compactitem}
} 

\noindent\walon{As previously discussed, if the masking operation used in these three masking strategies is completely random (i.e. the regions/patches to be masked are sampled randomly), there could exist the potential case where all the masked patches fall on either the foreground or the background objects thus leading to improper contrastiveness (e.g. forming a positive pair where one is fully background
while the other still contains most of the foreground). To this end, we propose to use the saliency information (i.e. the saliency map $M$ obtained by our localization network $f_{\xi}$ stemmed from SCDA technique) to guide the masking operation used in all our three masking strategies. Basically, the saliency map $M$ helps us to separate the foreground objects and the background, in which we perform random masking independently for foreground and background (i.e. distribute the masked patches more evenly to both foreground and background). In the following, we detail how we utilize saliency to create positive and (hard) negative samples for driving the contrastive self-supervised learning objective. 
}

Assume an input image $X$ is composed of $N$ patches and $\gamma$ denotes the ratio of $N$ patches that are identified as foreground by SCDA (where $N=U\times V$, i.e. the total number of elements in saliency map $M$ and there are $\gamma \cdot N$ patches be identified as foreground). Given a masking ratio $\alpha$, as we would like to have the masking performed separately for both foreground and background, for a positive sample, the ratio of the number of masked patches between foreground and background is $\gamma:(1-\gamma)$, i.e. there are $\alpha\cdot \gamma \cdot N$ patches randomly chosen to be masked in the foreground (respectively $\alpha \cdot (1-\gamma)\cdot N$ randomly-masked patches in the background). Noting that in our experiments $\alpha$ is drawn from a uniform distribution $\mathcal{U}(0.05, 0.25)$ unless otherwise stated;  When it comes to creating \textit{hard negative samples}, the masking is only applied on the foreground patches (i.e. the main objects are mostly masked to remove the salient/important information of such input image). We achieve so by drawing $\beta \sim \mathcal{U}(0.4,\, 0.7)$ in our experiments and randomly masking $\beta\cdot \gamma \cdot N$ foreground patches.

\begin{figure*}[ht!]
  \begin{minipage}[h]{0.765\textwidth} 
    \centering
    \includegraphics[width=1.0\textwidth]{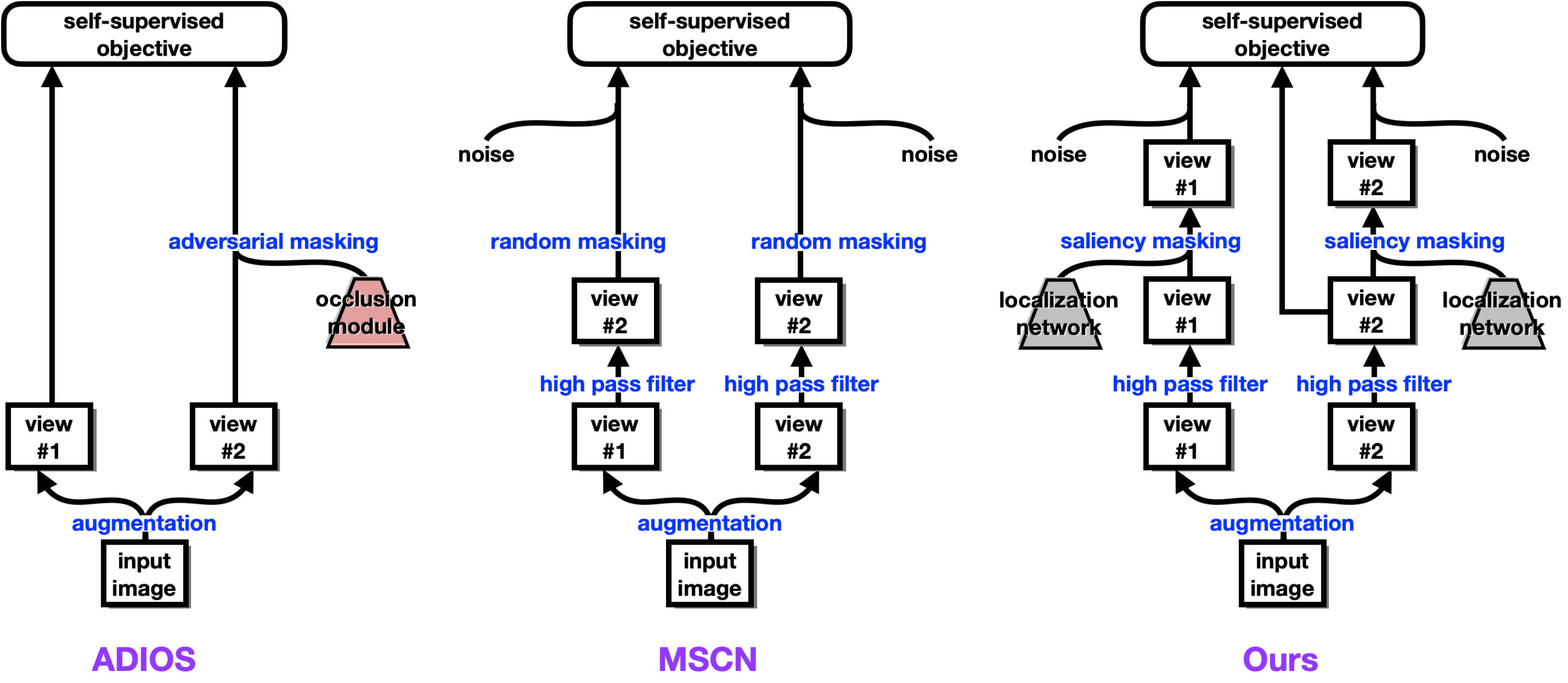} 
    \vspace{-1.5em}
    \caption{\walon{Comparison in terms of modelling among MSCN~\cite{jing2022masked}, ADIOS~\cite{shi2022adversarial}, and our proposed method. Please refer to the last paragraph of Section~\ref{sec:related} for more detailed descriptions.}}
    \label{fig:model_comparison} 
  \end{minipage}
  \hspace{0.025\columnwidth}
  \begin{minipage}[h]{0.215\textwidth} 
    \centering
    \includegraphics[width=1.0\textwidth]{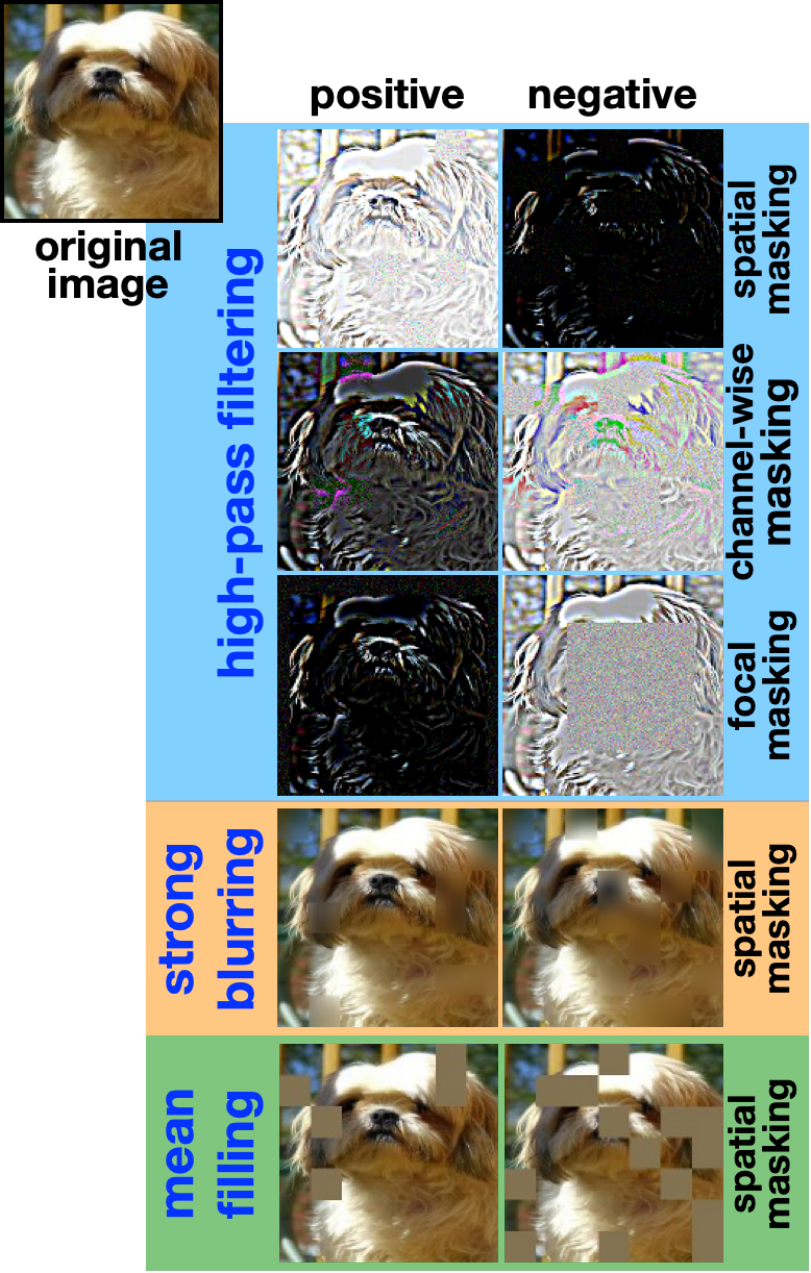}
    \vspace{-1.5em}
    \caption{\walon{Three masking strategies (cf. Section~\ref{Masking strategy}).}}
    \label{fig:masking_strategies} 
  \end{minipage}%
\vspace{-1.5em}
\end{figure*}

\walon{Since the saliency-guided masking operation described above is applied upon the spatial dimension, we also name it as \textbf{spatial masking} (to make analogy to the terminology defined in MSCN~\cite{jing2022masked}, but please note that ours particularly has the guidance of saliency), and such saliency-guided spatial masking operation is adopted for all our three masking strategies. Moreover, as the masking strategy of high-pass filtering is inspired by MSCN~\cite{jing2022masked}, we also include/extend two other masking operations used in MSCN to our high-pass filtering strategy: \textbf{channel-wise masking} where our saliency-guided spatial masking operation is applied individually for each of the RGB channels, and \textbf{focal masking} where random cropping is performed (noting that such focal masking does not involve any saliency guidance). 
Specifically, for focal masking, the region outside of $200\times 200$ (respectively region inside $130\times 130$) is cropped and replaced by Guassian noise to produce positive samples (respectively hard negative samples) is our experiments.
Furthermore, to be even more aligned with the original masking operation in MSCN~\cite{jing2022masked}, in our high-pass filtering strategy we will add random Gaussian noise to the masked sample regardless of which masking operation (i.e. spatial, channel-wise, and focal masking) is adopted (noting that for the strong blurring and mean filling masking strategies we do not apply such step). In Figure~\ref{fig:masking_strategies} we provide examples of positive and hard negative samples from all our three masking strategies.
}

\subsection{Learning scheme}\label{Overall design}
\vskip -0.1cm
\walon{Given the saliency-guided masking strategies introduced above, here we summarize the overall learning scheme for our including masking augmentation into contrastive self-supervised learning framework, where the feature extractor is based on ConvNet-backbone. Following the common scenario of contrastive SSL, two views of an input image $X$ are firstly produced by two different standard augmentations, where one view is denoted as the \textit{key view} while the other is denoted as \textit{query view}. Afterwards, we can apply saliency-guided masking operation (parameterized by $\gamma$ and $\alpha$) to the query view, where the resultant masked query view $X_q$ together with the key view $X_{k^+}$ form the positive pair; or, we can apply saliency-guided masking operation (now parameterized by $\gamma$ and $\beta$) upon the key view, which results to be the hard negative sample $X_{k^-}$ to the original key view $X_{k^+}$. Noting that the masked query view $X_q$ together with the views of any other image different from $X$ (denoted as $X_{\neg}$) naturally forms the negative pairs. With denoting the feature representation of $X_q$ extracted by the feature encoder as $z_q$ (analogously $z_{k^+}$ for $X_{k^+}$, $z_{k^-}$ for $X_{k^-}$, and $z_{\neg}$ for $X_{\neg}$), our contrastive objective $\mathcal{L}_{nce}$ is built to pull closer the features of positive pair while pushing away the feature of negative pair:
\begin{equation}
\label{eq:L_nce}
\mathcal{L}_{nce}=-\log\frac{\exp(z_q^\top z_{k^+}/\tau)}{\displaystyle\sum_{X_{\neg}}\exp(z_q^\top z_{\neg}/\tau) + \exp(\rho z_q^\top z_{k^-}/\tau)}
\end{equation}
where $\tau$ is a temperature parameter and $\rho$ is the penalty ratio for hard negative samples, in which our $\mathcal{L}_{nce}$ is stemmed from InfoNCE loss~\cite{oord2018representation} and analogous to the one in~\cite{ge2021robust}.
}

\walon{It is worth noting that while constructing the positive pairs, the key view $X_{k^+}$ does not undergo any saliency-guided masking operation. Such design is motivated from 1) empirical findings that the masked views show higher variance than the views produced by standard augmentations (in which we provide the corresponding study in Section~\ref{sec:ablation}), and 2) the statement made in~\cite{wang2022importance} that having higher variance in the query branch than the key branch yields better results in the siamese network, where the benefit of such design to model training is demonstrated in our experiments provided in Table~\ref{tab:mask-diff-br}.
}

%% file: exp.tex
\section{Experimental Results}\label{sec:exp}
\vskip -0.1cm
We compare our proposed method with two state-of-the-art baselines including masking operations into the ConvNet-based SSL, i.e. MSCN~\cite{jing2022masked} and ADIOS~\cite{shi2022adversarial}, as illustrated in Figure \ref{fig:model_comparison}.
\walon{Following ADIOS~\cite{shi2022adversarial}, we adopt the ImageNet-100 dataset \cite{ILSVRC15} as the basis to conduct our experiments (i.e., being used to perform contrastive self-supervised learning for training the feature encoder), while we choose MoCov2~\cite{chen2020improved} and SimCLR~\cite{chen2020simple} to be our experimental bed of contrastive SSL frameworks. Basically, ImageNet-100 dataset contains 100 ImageNet classes, and is composed of 1300 training images and 50 validation images for each class. We use ResNet-50 as our ConvNet-backbone for the feature encoder for both contrastive SSL frameworks. For MoCov2, we set the batch size to 128 and the learning rate to 0.015, and use SGD \cite{ruder2016overview} as the optimizer; While for SimCLR, we set the batch size to 256 and the learning rate to 0.3, and use LARS \cite{you2017large} as the optimizer. The contrastive self-supervised pretraining is run on a 4-GPU machine for 200 epochs, including 10 epochs of warm-up and using a cosine learning rate scheduler.
}
\subsection{ImageNet-100 Classification}
\vskip -0.1cm
\begin{table}[t]{}
\begin{center}
\begin{tabular}{l|c}
Method & Linear Evaluation \\
\Xhline{1.5pt}
MoCov2 ~\cite{chen2020improved} & 68.22 \\
+ MSCN~\cite{jing2022masked} & 70.28 \\
+ ADIOS~\cite{shi2022adversarial} & 62.76 \\
+ OURS (High-pass filtering) & \textbf{73.8} \\
+ OURS (Strong blurring) & 72.50 \\
+ OURS (Mean filling) & 70.84 \\
\hline
SimCLR~\cite{chen2020simple} & 69.77 \\
+ MSCN~\cite{jing2022masked} & 77.18 \\
+ ADIOS~\cite{shi2022adversarial} & 71.12 \\
+ OURS (High-pass filtering) & \textbf{77.9} \\
+ OURS (Strong blurring) & 77.78 \\ 
+ OURS (Mean filling) & 77.36 \\ 
\end{tabular}
\end{center}
\vspace{-1.4em}
\caption{Linear evaluation results on ImageNet-100 classification task, where MoCov2 or SimCLR are used as the contrastive SSL framework for pretraining the feature encoder. The best results are marked in bold.}
\label{tab:imagenet100}
\vskip -0.25in
\end{table}

\walon{Once the feature encoders are pretrained via adopting our proposed method (with three different masking strategies) and baselines (i.e. MSCN and ADIOS) in the contrastive SSL frameworks (i.e. MoCov2 and SimCLR), we now evaluate their performances on various downstream tasks. Firstly, we experiment on the downstream task of classification based on ImageNet-100 dataset, where a linear classifier is supervisedly trained while the feature encoder is kept fixed/frozen. Please note again that, as described in Section~\ref{Masking strategy}, for the feature encoder pretrained by using the high-pass filtering masking strategy, its input in the downstream task should still be firstly gone through the high-pass filter (in which such requirement is also applied to MSCN). The evaluation results on ImageNet-100 classification are summarized in Table~\ref{tab:imagenet100}, where our proposed method (regardless of adopting any saliency-guided masking strategies in both SSL frameworks) consistently achieves superior performance comparing to MoCov2 baseline (i.e. no masking augmentation is applied), MoCov2+MSCN, and MoCov2+ADIOS, where the similar trend is also observable while using SimCLR as the contrastive SSL framework, thus verifying the contribution and the efficacy of our proposed saliency-guided masking methods. It is worth noting that, although our high-pass filtering masking strategies shares quite some common designs as MSCN, our explicit introduction of saliency guidance contributes to the resultant improvement of our proposed method, e.g. MoCov2+OURS (High-pass filtering) versus MoCov2+MSCN and SimCLR+OURS (High-pass filtering) versus SimCLR+MSCN in Table~\ref{tab:imagenet100}.}

\subsection{Transfer Learning}
\vskip -0.1cm
\walon{As one of the important goals of SSL is to obtain the feature encoder with better generalizability such that the encoder can be easily adapted to various tasks or datasets with little amount of labeled data, we thus further conduct experiments on different downstream tasks or datasets for better assessing the generality of the features learned by various methods. Here we take MoCov2 as the main experimental bed of SSL framework, and we adopt high-pass filtering masking strategy to present our proposed method for making comparison with the baselines (while the results of adopting strong blurring and mean filling masking strategies in our proposed method are provided in the Appendix). }

\begin{table*}[ht]{}
\begin{center}
\begin{tabular}{l|ccc|ccc|ccc}
    \multirowcell{2}{Method} & \multicolumn{3}{c|}{VOC07+12 detection} & \multicolumn{3}{c|}{COCO detection} & \multicolumn{3}{c}{COCO instance segmentation} \\
                             & $AP_{all}$ & $AP_{50}$ & $AP_{75}$ & $AP^{bb}_{all}$ & $AP^{bb}_{50}$ & $AP^{bb}S_{75}$ & $AP^{mk}_{all} $ & $AP^{mk}_{50}$ & $AP^{mk}_{75}$ \\
\Xhline{1.5pt}
    MoCov2     & 50.27 & 76.68 & 54.76 & 38.52 & 57.62 & 41.67 & 33.75 & 54.70 & 35.86 \\
    \hline
    +  MSCN  & 50.27 & 76.99 & 54.70 & 38.80 & 58.09 & 42.20 & 33.89 & 54.78 & \textbf{36.36}\\
    +  ADIOS & 45.85 & 73.44 & 48.45 & 38.12 & 57.38 & 41.29 & 33.38 & 54.25 & 35.63\\
    \hline
    +  OURS (High-pass filtering) & \textbf{50.89} & \textbf{77.66} & \textbf{55.44} & \textbf{39.16} & \textbf{58.62} & \textbf{42.45} & \textbf{34.22} & \textbf{55.28} & 36.30 \\
\end{tabular}
\end{center}
\vspace{-1.2em}
\caption{Transfer learning results on VOC07+12 and COCO detection tasks, and COCO instance segmentation task. Performances in terms of $AP_{all}$, $AP_{50}$ and $AP_{75}$ metrics are reported, and the best results are marked in bold.}
\label{tab:det_seg}
\vskip -0.2in
\end{table*}

\begin{table}[t]{}
\vskip 0.05in
\begin{center}
\begin{tabular}{l|cc}
    \multirowcell{1}{Method} & Caltech-101 & Flowers-102\\
\Xhline{1.5pt}
    MoCov2   & 81.87   & 88.39  \\
    \hline
    +  MSCN~\cite{jing2022masked}  & 84.13   & 90.10  \\
    +  ADIOS~\cite{shi2022adversarial} & 79.83   & 88.39   \\
    \hline
    +  OURS (High-pass filtering)  & \textbf{84.91} & \textbf{90.95}  \\
\end{tabular}
\end{center}
\vspace{-1.4em}
\caption{Transfer learning results on Caltech-101 and Flowers-102 classification tasks. 
}
\label{tab:transfer_classification}
\vskip -0.2in
\end{table}

\vskip -0.1in
\noindent\textbf{Image classification on different datasets.} 
\walon{Here we experiment on the classification downstream task based on two widely used benchmarks, i.e. Caltech-101 \cite{fei2004learning} and Flowers-102 \cite{nilsback2008automated}, which are different from the one used for feature encoder pretraining (i.e. ImageNet-100). Again, we keep the pretrained feature encoder fixed and only train the linear classifier when learning the downstream task. The experimental results are reported in Table~\ref{tab:transfer_classification}, where our proposed method outperforms both MSCN and ADIOS baselines on both Caltech-101 and Flowers-102 datasets.}

\noindent\textbf{Object detection and instance segmentation.} 
\walon{Now we turn to different downstream tasks on object detection and instance segmentation, where the former is conducted on VOC07+12 \cite{everingham2010pascal} and COCO \cite{lin2014microsoft} datasets while the latter is conducted on the COCO dataset. With keeping the feature encoder fixed and only supervisedly training the detection or segmentation heads (where for VOC07+12 dataset we adopt the Faster-RCNN \cite{ren2015faster} model with a C4 backbone which finetuned for 24k iterations, while for COCO dataset we adopt Mask R-CNN \cite{he2017mask} model with C4 backbone which is finetuned for 180K iterations, following the same experimental setting as in the original MoCov2 paper), the experimental results are summarized in Table~\ref{tab:det_seg}. From the results we can again observe the consistent outperformance with respect to both MSCN and ADIOS baselines.}

\walon{The transfer learning results provided in Table \ref{tab:transfer_classification} and Table \ref{tab:det_seg} support that our method can effectively learn general-purpose features which can be transferred across different downstream tasks or datasets, providing a promising finding for future research in the field of self-supervised learning.}

\subsection{Ablation Study}\label{sec:ablation}
\vskip -0.1cm
\walon{Our ablation studies are carried out on TinyImageNet \cite{Le2015TinyIV}, a well-known subset of the ImageNet-1K dataset \cite{ILSVRC15}. This dataset consists of 200 classes, comprising 500 training images and 50 validation images in each class. Compared to ImageNet-100, TinyImageNet contains more classes but fewer training images per class, which is even more challenging for the model to learn. As a result, we take it to better verify the contribution for each of our designs. We employ MoCov2 as our SSL framework with using ResNet-50 as the backbone for feature encoder, which is pretrained for 200 epochs. The evaluation is conducted by the downstream task of classification on TinyImageNet.} If not mentioned, the positive masking ratio is set to 15\% of the whole image, and the hard negative masking ratio is set to 40\%-70\% salient patches.

\begin{table}[t]{}
\vskip 0.05in
\begin{center}
\begin{tabular}{c|cc|c}
  Setting &  Saliency & Mask non-salient & Top1 \\
\Xhline{1.5pt}
\multirowcell{3}{High-pass \\ filtering} & \xmark & \cmark & 56.15 \\
                                      & \cmark & \cmark & \textbf{56.60} \\
                                      & \cmark & \xmark & 55.87 \\
\hline
\multirowcell{3}{Strong \\ blurring} & \xmark & \cmark & 55.33 \\ 
                                      & \cmark & \cmark & \textbf{56.39} \\
                                      & \cmark & \xmark & 54.37 \\
\hline
\multirowcell{3}{Mean \\ filling} & \xmark & \cmark & 55.93 \\ 
                             & \cmark & \cmark & \textbf{55.94} \\ 
                             & \cmark & \xmark & 55.59 \\ 
\end{tabular}
\end{center}
\vspace{-1.4em}
\caption{We conduct three different experiments on TinyImageNet classification task to justify the importance of saliency guidance.}
\label{tab:ab-salient}
\vskip -0.2in
\end{table}

\noindent\textbf{Impact of Saliency.}
\walon{Saliency is the soul of our work and has always been mentioned in this paper. We compare three settings: pure random masking (i.e. MSCN), masking with saliency constraint, and masking on salient patches only, and the results are reported in Table \ref{tab:ab-salient}. Our findings indicate that adding a saliency constraint can benefit our training, but only if we distribute the masked patches to the foreground and background patches. Surprisingly, masking totally on salient patches results in poor performance, possibly due to the reason that such processing causes the model to rely more heavily on the background, which can be detrimental to the overall performance.}

\begin{table*}[t]{}
\begin{center}
\begin{tabular}{c|ccc|ccc}
    \multirowcell{2}{Pretrained dataset} & ImageNet-100 & Caltech-101 & Flowers-102 & \multicolumn{3}{c}{VOC07+12 det} \\
                                          & \multicolumn{3}{c|}{Top1} & $\text{AP}_{\text{all}}$ & $\text{AP}_{50}$ & $\text{AP}_{75}$ \\
\Xhline{1.5pt}
    ImageNet-1K & 73.80 & 84.91 & 90.95 & 50.89 & 77.66 & 55.44  \\
    COCO & 73.78 & 85.68 & 90.83 & 50.22 & 77.41 & 54.28 \\
\end{tabular}
\end{center}
\vspace{-1.4em}
\caption{\walon{We compare the results of using localization networks pretrained on the ImageNet-1K and COCO dataset, where the localization network is to produce the saliency maps for guiding our masking operations. Similar performances produced by using these two localization networks demonstrate that our proposed method is insensitive to the selection of localization network for performing saliency computation, once it provides feasible localization results.}
}
\label{tab:coco_localize_work}
\vskip -0.2in
\end{table*}

\begin{table}[t]{}
\vskip 0.05in
\begin{center}
\begin{tabular}{c|c|c}
Setting & Mask branch & Top1 \\
\Xhline{1.5pt}
Baseline MoCov2 & \xmark & 56.00 \\
\hline
\multirowcell{3}{High-pass filtering} &  key & 52.25 \\
                                             &  both & 56.29 \\
                                             &  query & \textbf{58.19} \\
\hline
\multirowcell{3}{Strong blurring} & key & 51.06  \\
                                    & both & 56.83  \\
                                    & query & \textbf{58.28} \\
\hline
\multirowcell{3}{Mean filling} & key & 47.53 \\
                                    & both & 56.86 \\ 
                                    & query &  \textbf{58.34}  \\ 
\end{tabular}
\end{center}
\vspace{-1.4em}
\caption{
Comparison of masking on different branches of SSL framework.}

\label{tab:mask-diff-br}
\vskip -0.18in
\end{table}

\begin{table}[ht]{}
\begin{center}
\begin{tabular}{l|c}
    \multirowcell{1}{Augmentation}  & Variance (1e-3)\\
\Xhline{1.5pt}
    Standard     & 6.196  \\
    \hline
    +  High-pass filtering masking  & 10.952 \\
    \hspace{9pt} High-pass filtering w/o masking & 8.67 \\ 
    +  Strong blurring masking   & 7.744  \\
    +  Mean filling masking        & 7.776  \\
\end{tabular}
\end{center}
\vspace{-1.4em}
\caption{
Variance of the representations b/w standard augmentation and our saliency masking (refer to \cite{wang2022importance} for variance calculation, based on TinyImageNet validation set).}
\label{tab:variance}
\vskip -0.2in
\end{table}


\noindent\textbf{Masking Controls Variance.}
\walon{Our approach involves masking only the query branch when constructing positive pairs.
To investigate the effects of masking different branches on performance, we conduct a study consisting of three experiments, which are masking the key branch only, masking both branches, and masking the query branch only. The results of this study, as presented in Table \ref{tab:mask-diff-br}, indicate that masking the query branch only performs the best, and masking the key branch only performs worse than the baseline MoCov2. Such results are aligned with \cite{wang2022importance}, which suggests that maintaining a lower variance in the key branch than in the query branch during pretraining can be beneficial, and not the other way around. Moreover, we hypothesize that masking can also influence variance such that manipulating it through masking can lead to better results. To further support our claim, we conduct an experiment comparing the variance of standard data augmentation with standard data augmentation combined with our saliency masking for all of our settings, where the results are presented in Table \ref{tab:variance}. We can observe that standard data augmentation combined with our saliency masking leads to a higher variance than standard data augmentation for all our settings.}

\begin{table}[t]{}
\vskip 0.15in
\begin{center}
\begin{tabular}{c|c|c|c}
    Setting & Positive mask & Negative mask & Top1 \\
\Xhline{1.5pt}
\multirowcell{2}{High-pass \\filtering} & \cmark & \xmark & 55.25 \\
                                      & \cmark & \cmark & \textbf{56.26} \\
\hline
\multirowcell{2}{Strong \\ blurring} & \cmark & \xmark & 55.52 \\
                         & \cmark & \cmark & \textbf{56.83} \\
\hline
\multirowcell{2}{Mean \\ filling} & \cmark & \xmark & 55.18 \\
                             & \cmark & \cmark & \textbf{56.21} \\
\end{tabular}
\end{center}
\vspace{-2em}
\caption{
Efficacy of our proposed hard negative samples.}
\label{tab:ab-hard-neg}
\vskip -0.25in
\end{table}

\noindent\textbf{Impact of Hard Negative Samples.}
\walon{We conduct a study to verify our designs of creating hard negative samples by masking large portion of salient patches (40\%-70\%) of the key view.}
Table \ref{tab:ab-hard-neg} compares the results of being with or without our proposed hard negative samples, showing the benefit brought by our designs. By masking more salient patches, the remaining part of image has more background than foreground. We deem this view as negative one, though it comes from the same sample image as the positive view. The model might be confused if it is biased to the background when the similarity between the hard negative view and the query view is too high. In turn, the model with our hard negative samples will focus on learning more from the foreground thus boosting the performance.


\noindent\textbf{Impact of Localization Networks Pretrained on Different Datasets.}
\walon{As we adopt ImageNet-pretained classfication model as the basis for SCDA to build our localization network (for producing saliency maps), there could exist potential concern if we take any advantage later in the SSL pretraining stage than other baselines. To resolve such potential concern, here we conduct a study to use another model, the ResNet-50 backbone from a Faster R-CNN detection model pretrained on the COCO dataset, as the basis for SCDA. With using our high-pass filtering masking strategy guided by the saliency maps respectively from different localization networks in MoCov2 SSL framework to training the feature encoders, the experimental results upon various downstream tasks and datasets (following the same settings as previous experiments) are summarized in Table~\ref{tab:coco_localize_work}. We are able to observe that both localization networks result in similar performances, thus verifying that our method is not sensitive to the selection of localization network once it is able to provide reasonable localization capability.}

%% file: conclusion.tex
\section{Conclusion}
\vskip -0.1cm
\walon{We propose a salient masking augmentation method for contrastive self-supervised learning with a ConvNet as its backbone. Compared to randomly masking patches of the input image, our salient masking provides more semantically meaningful masks while its efficacy is well verified in our ablation study. Besides masked positive samples, we further introduce a simple way to create hard negative samples according to three different masking strategies, which further improve the capability of training the feature encoder. The extensive experimental results demonstrate the effectiveness and superiority of our proposed method.}

%% file: supp.tex
\newcommand{\best}[1]{\textbf{\textcolor{orange}{#1}}}
\newcommand{\secondbest}[1]{\textbf{\textcolor{blue}{#1}}}

\definecolor{col1}{RGB}{232, 161, 148}
\definecolor{col2}{RGB}{148, 187, 232}

\appendix
\section{Appendix}
In this appendix, we firstly provide the summarization of our contributions with respect to our two main baselines MSCN~\cite{jing2022masked} and ADIOS~\cite{shi2022adversarial} (cf. Section~\ref{mscn} and Section~\ref{adios} respectively) as well as our contribution in terms of saliency masking (cf. Section \ref{saliency}). Furthermore, we show the efficacy of our three different masking strategies (i.e., high-pass filtering, strong blurring, and mean filling) with more experiments as well as discuss the computational cost: In Section \ref{exp}, we provide detailed experimental setups and conduct various downstream tasks (i.e., classification, object detection, and semantic segmentation) in different contrastive SSL frameworks (i.e., MoCov2~\cite{chen2020improved} and SimCLR~\cite{chen2020simple}); While in Section \ref{time_cost}, we compare the computational cost of three different masking strategies with  MSCN~\cite{jing2022masked} and ADIOS~\cite{shi2022adversarial}.

\subsection{Emphasis upon our contribution compared to baseline MSCN~\cite{jing2022masked}}\label{mscn}
Here we would like to emphasize again that our contributions stand out from the ones of MSCN~\cite{jing2022masked} as it includes: 1) \textit{\underline{Saliency masking} \underline{with various masking strategies}} (their benefits are shown in Table 4 and 1 of the main manuscript), in which MSCN does not adopt saliency-guided masking but applies random masking, and its masking strategy based on high-pass filtering constrains the setting of downstream tasks (since the input for the downstream tasks needs to be firstly high-pass-filtered as well, i.e. having the prior knowledge upon how the pre-training of feature extractor is done, c.f. lines 360-372 in our main manuscript). Our proposed strong blurring and mean-filling masking strategies are novel and practical as they do not have such constraints, thus being more flexible; 2) Based on the explicit analysis of \textit{\underline{variance manipulation}}, our proposed method applies masking solely on the query branch of the siamese framework and is shown to consistently improve the performance for all masking strategies (c.f. Table 6 of the main manuscript); 3) Generating the \textit{\underline{hard negative samples}} easily by masking only the foreground patches with the help of saliency (cf. Table 8 of the main manuscript for the improvement based from such design).

\subsection{Emphasis upon our contribution compared to baseline ADIOS~\cite{shi2022adversarial}}\label{adios}

We would like to emphasize that our contributions stand out from the ones of ADIOS~\cite{shi2022adversarial} as it includes: 1) \textit{Efficiency in obtaining (partially) semantic masks}. 
While both our proposed method and ADIOS employ a localization network to address the ``where to mask'' issue, our approach achieves a more favorable trade-off between obtaining (partially) semantic masks and computational effort. Notably, the localization network we utilize remains frozen during feature extractor training, whereas ADIOS requires joint training of the localization network (UNet-based segmentation model) alongside the feature extractor; 2) \textit{Variance manipulation in single branch}. ADIOS masks a \textit{single view} while both views (i.e. masked and unmasked) will go through both query and key branches (as indicated in their source code). In comparison, our design shows that incorporating variance manipulation through masking only the query branch has a positive impact on the Siamese network. Our method differs from ADIOS in terms of both operation and motivation (i.e. variance manipulation). Further details of our investigation and discussion can be found in lines 744-807, and while corresponding ablation studies can be found in Tables 6 and 7.

\subsection{Our contribution in saliency masking}\label{saliency}
As described in lines 101-113 in our main manuscript, and we would like to clarify again here: most existing studies of adopting masking operations (together with self-reconstruction objective) to realize self-supervised learning are based on the \textit{transformer backbone} thanks to the tokenized input (where the masking is simply to block out some tokens), and the prior works (e.g. SemMAE~\cite{li2022semmae}, MST~\cite{li2021mst}, BEiT~\cite{bao2021beit}, iBOT~\cite{zhou2021ibot}, MAE~\cite{he2022masked}, and SimMIM~\cite{xie2022simmim}) are designed for transformers as well. In contrast, we aim to apply masking for \textit{convolutional neural networks}, which is actually nontrivial due to the unwanted edges caused by masking (and that is exactly why MSCN~\cite{jing2022masked} needs to introduce the high-pass filtering at first). Moreover, even there exists some transformer-based prior works adopting the saliency operations as well, the ways of their applying saliency masking are also different from ours: For instance, SemMAE~\cite{li2022semmae} requires a two-stage training process to determine where to apply the mask, while our approach achieves the same goal with a single feature extractor and end-to-end training; MST~\cite{li2021mst} also aims to avoid masking important objects, while our method of explicitly distributing masked patches across foreground and background empirically leads to better performance.

\subsection{More Experimental Results}\label{exp}
In this section, we provide the results for all the downstream tasks with three different masking strategies (i.e., high-pass filtering, strong blurring, and mean filling) and baselines (i.e., MSCN~\cite{jing2022masked} and ADIOS~\cite{shi2022adversarial}) based on two contrastive SSL frameworks (i.e., MoCov2~\cite{chen2020improved} and SimCLR~\cite{chen2020simple}). In the pretraining stage, we train the feature encoder (under MoCov2 and SimCLR frameworks) with using ResNet-50 as the backbone on the ImageNet-100~\cite{ILSVRC15} dataset for 200 epochs. We conduct experiments on three datasets (i.e., ImageNet-100, Caltech-101~\cite{fei2004learning}, and Flower-102~\cite{nilsback2008automated}) for downstream classification tasks and supervisedly train a linear classifier while the feature encoder is kept fixed/frozen for 100 epochs. We conduct experiments on VOC07+12~\cite{everingham2010pascal} and COCO~\cite{lin2014microsoft} datasets for downstream detection tasks, where the COCO dataset is also used for the downstream instance segmentation task. For the VOC07+12 dataset, we adopt the Faster R-CNN~\cite{ren2015faster} model with C4 backbone which is finetuned for 24k iterations; while for the COCO dataset, we adopt the Mask R-CNN~\cite{he2017mask} model with C4 backbone which is finetuned for 180k iterations (using 1$\times$ learning rate schedule).

\textbf{MoCov2 Results.} 
\textcolor{cyan}{First of all, please note that most of the results based on MoCov2 framework have been provided in our main paper, here we particularly include them again for the purpose of having better and more complete overview.}
For MoCov2, we set the batch size to 128 and the base learning rate to 0.015 and use SGD~\cite{ruder2016overview} as the optimizer during pretraining. When training the linear classifier, we set the base learning rate to 30.0 and adopt a learning rate schedule that decreases the learning rate by 0.1 at epochs 60 and 80. All MoCov2's downstream classification results are reported in the upper half of Table \ref{tab:downstream_classification}, while all the downstream detection and instance segmentation results are reported in the upper half of Table \ref{tab:det_seg_supp}. Our method outperforms the fundamental contrastive SSL framework (i.e., MoCov2, which has no masking involved) and two baselines (i.e., MSCN and ADIOS) in all the downstream tasks. 

\begin{table*}[t]{}
\vskip 0.05in
\begin{center}
\begin{tabular}{l|cccc}
    \multirowcell{1}{Method} & ImageNet-100 & Caltech-101 & Flowers-102\\
\Xhline{1.5pt}
    Supervised & 82.72 & 21.99 & 20.29 \\
    \hline
    \hline
    MoCov2 & 68.22  & 81.87   & 88.39  \\
    \hline
    +  MSCN~\cite{jing2022masked} & 70.28 & \secondbest{84.13}   & 90.10  \\
    +  ADIOS~\cite{shi2022adversarial} & 62.76 & 79.83   & 88.39   \\
    \hline
    +  OURS (High-pass filtering) & \best{73.80}  & \best{84.91} & \best{90.95}  \\
    +  OURS (Strong blurring) & \secondbest{72.50} & 83.95 & 90.59 \\
    +  OURS (Mean filling) & 70.84 & 82.68 & \secondbest{90.83} \\
    \hline
    \hline
    SimCLR & 69.77 &  78.20  & 85.21  \\
    \hline
    +  MSCN~\cite{jing2022masked} & 77.18 & \secondbest{86.99} & \secondbest{91.08}  \\
    +  ADIOS~\cite{shi2022adversarial} & 71.12 & 81.96 & 87.53  \\
    \hline
    +  OURS (High-pass filtering) & \best{77.90}  & \best{87.04} & 90.71  \\
    +  OURS (Strong blurring) & \secondbest{77.78} & 83.41 & \best{91.93}  \\
    +  OURS (Mean filling) & 77.36 & 83.55 & 90.83 \\
\end{tabular}
\end{center}
\vspace{-1em}
\caption{Linear evaluation results on ImageNet-100, Caltech-101 and Flowers-102. The best and second-best results on each dataset with different constrastive SSL frameworks (i.e., MoCov2, SimCLR) are marked in orange and blue respectively.}
\label{tab:downstream_classification}
\end{table*}

\begin{table*}[ht]{}
\vskip 0.05in
\begin{center}
\begin{tabular}{l|ccc|ccc|ccc}
    \multirowcell{2}{Method} & \multicolumn{3}{c|}{VOC07+12 detection} & \multicolumn{3}{c|}{COCO detection} & \multicolumn{3}{c}{COCO instance segmentation} \\
                             & $AP_{all}$ & $AP_{50}$ & $AP_{75}$ & $AP^{bb}_{all}$ & $AP^{bb}_{50}$ & $AP^{bb}_{75}$ & $AP^{mk}_{all} $ & $AP^{mk}_{50}$ & $AP^{mk}_{75}$ \\
\Xhline{1.5pt}
    Supervised & 44.30 & 73.47 & 46.50 & 37.84 & 57.09 & 40.67 & 33.14 & 53.95 & 35.31 \\
    \hline
    \hline
    MoCov2     & 50.27 & 76.68 & 54.76 & 38.52 & 57.62 & 41.67 & 33.75 & 54.70 & 35.86 \\
    \hline
    +  MSCN  & 50.27 & 76.99 & 54.70 & 38.80 & 58.09 & \secondbest{42.20} & 33.89 & 54.78 & \secondbest{36.36}\\
    +  ADIOS & 45.85 & 73.44 & 48.45 & 38.12 & 57.38 & 41.29 & 33.38 & 54.25 & 35.63\\
    \hline
    +  OURS (High-pass filtering) & \best{50.89} & \best{77.66} & \best{55.44} & \best{39.16} & \best{58.62} & \best{42.45} & \best{34.22} & \best{55.28} & 36.30 \\
    +  OURS (Strong blurring) & \secondbest{50.76} & \secondbest{77.29} & 54.75 & 38.90 & \secondbest{58.13} & 42.11 & \secondbest{33.93} & 54.77 & \best{36.53} \\
    +  OURS (Mean filling) & 50.59 & 76.97 & \secondbest{55.30} & \secondbest{38.93} & 58.08 & 42.17  & 33.92 & \secondbest{54.86} & 36.27 \\
    \hline
    \hline
    SimCLR   & 40.34 & 69.86 & 40.96 & 36.30 & 55.55 & 38.80 & 31.99  & 52.28  & 33.80  \\
    +  MSCN  & 43.50 & 73.18 & \best{45.04}  & 37.88 & 57.44 & 40.68 & 33.36 & 54.15 & 35.57 \\
    +  ADIOS & \best{43.83} & \secondbest{73.42} & \secondbest{45.01} & \best{38.76} & \best{58.35} & \best{41.96} & \best{33.94} & \best{54.96} & \best{36.23} \\
    \hline
    +  OURS (High-pass filtering) & \secondbest{43.76} & \best{73.43} & 44.90  & \secondbest{38.45} & \secondbest{57.79} & \secondbest{41.58} & \secondbest{33.90} & \secondbest{54.70} & \secondbest{35.93} \\
    +  OURS (Strong blurring) & 43.20 & 73.15 & 44.27 & 37.44 & 56.80 & 39.96 & 32.92 & 53.73 & 35.00  \\
    +  OURS (Mean filling) & 43.20  & 72.54 & 44.79 & 37.27 & 56.46 & 40.10  & 32.68 &  53.35 & 34.54 \\
\end{tabular}
\end{center}
\vspace{-1em}
\caption{Transfer learning results on VOC07+12 and COCO detection tasks, and COCO instance segmentation task. Performances in terms of $AP_{all}$, $AP_{50}$ and $AP_{75}$ metrics are reported, and the best and second-best results on each task of different contrastive SSL frameworks (i.e., MoCov2, SimCLR) are marked in orange and blue respectively.}
\label{tab:det_seg_supp}
\end{table*}

\textbf{SimCLR Results.}
For SimCLR, we set the batch size to 256, the base learning rate to 0.3, and use LARS~\cite{you2017large} as the optimizer during pretraining.  When training the linear classifier, we set the batch size to 256, the base learning rate to 1.0, and adopt a cosine learning rate schedule. All the SimCLR's downstream classification results are reported in the lower half of Table \ref{tab:downstream_classification}, while all the downstream detection and instance segmentation results are reported in the lower half of Table \ref{tab:det_seg_supp}.
Our method outperforms the fundamental contrastive SSL framework (i.e., SimCLR, which has no masking involved) and two baselines (i.e., MSCN and ADIOS) in all the classification tasks; but ADIOS slightly outperforms our method in the detection and instance segmentation tasks, where we attribute this to two reasons. Firstly, according to ablation studies conducted in~\cite{wang2022importance}, manipulating variance across branches in symmetric encoders (i.e. SimCLR) does not improve as much as that in asymmetric encoders (i.e., MoCov2), limiting improvement in our three masking strategies. Secondly, more detailed semantically meaningful masks of ADIOS are learnt in its pretraining stage, which yield better performance for the downstream detection and instance segmentation tasks (as both detection and instance segmentation can be seen as more detailed recognition tasks than classification). However, noting that an occlusion module needs to be trained jointly with the main SSL objective to learn such masks for ADIOS (thus being believed to require more computational efforts). In contrast, our saliency masking utilizes a pretrained localization network before masking (where the resultant masks are less detailed than the ones in ADIOS but no additional joint learning is required) and still contributes to the comparable results with ADIOS. 

\textbf{Supervised Baseline Results.}
To establish a solid foundation, we create a supervised baseline. In this baseline, we train an image classification model using ResNet-50 as the feature extractor. Our training setup involves using a batch size of 256, a base learning rate of 0.1, and a learning rate decay of 10 every 30 epochs, and employing SGD as our optimizer when training on the ImageNet-100 dataset. For downstream classification tasks involving Caltech-101 and Flowers-102, we follow our SSL approaches. In these cases, we kept the ResNet-50 feature which is trained on ImageNet-100 fixed, and train a linear classifier with hyperparameters similar to our SSL approaches. All classification tasks undergo 100 epochs of training. Regarding downstream detection and instance segmentation tasks, we utilize settings similar to those used in our SSL methods. The top row of Table \ref{tab:downstream_classification} presents the results for the supervised baseline classification, while the top row of Table \ref{tab:det_seg_supp} showcases the results for downstream detection and instance segmentation. Despite achieving the highest accuracy in ImageNet-100 classification, the supervised baseline exhibits the poorest transferability. Both transfer classification tasks achieve only 20\% accuracy, a result we attribute to the distribution differences between the ImageNet-100 and Caltech-101/Flowers-102 datasets.

\begin{table*}[ht]{}
\vskip 0.05in
\begin{center}
\begin{tabular}{l|cccc|ccc}
    Method & \cellcolor{col1}Abs Rel & \cellcolor{col1}Sq Rel & \cellcolor{col1}RMSE  & \cellcolor{col1}RMSE log & \cellcolor{col2}$\delta < 1.25 $ & \cellcolor{col2}$\delta < 1.25^{2}$ & \cellcolor{col2}$\delta < 1.25^{3}$\\
\Xhline{1.5pt}
    Monodepth2 \cite{monodepth2} & \best{0.132} & \secondbest{1.053} & \best{5.159} & \secondbest{0.211} & \best{0.846} & \secondbest{0.949} & 0.976 \\
    \hline
    \hline
    MoCov2     & 0.140 & 1.112 & 5.430 & 0.218 & 0.826 & 0.943 & \secondbest{0.977} \\
    +  MSCN  & 0.140 & 1.104 & 5.416 & 0.219 & 0.826 & 0.944 & \secondbest{0.977} \\
    +  ADIOS & 0.139 & 1.080 & 5.355 & 0.217 & 0.830 & 0.946 & 0.976 \\
    \hline
    +  OURS (High-pass filtering) & 0.138 & 1.074 & 5.331 & 0.216 & 0.828 & 0.945 & \secondbest{0.977} \\
    +  OURS (Strong blurring) & \secondbest{0.135} & 1.098 & 5.357 & 0.214 & 0.838 & 0.947 & \secondbest{0.977} \\
    +  OURS (Mean filling) & \best{0.132} & \best{1.043} & \secondbest{5.263} & \best{0.210} & \secondbest{0.842} & \best{0.950} & \best{0.979} \\
\end{tabular}
\end{center}
\vspace{-1em}
\caption{Transfer depth estimation results wih modified Monodepth2 through monocular training on KITTI 2015 \cite{Geiger2012CVPR} utilizing the Eigen split. Metrics presented in red cells denote that lower values are preferred, whereas those in blue cells suggest that higher values are desirable. The best results are marked in orange, while the second-best results are marked in blue.}
\label{tab:depth_estimation}
\end{table*}

\textbf{Monocular Depth Estimation Downstream Task Results}
In addition to the commonly addressed downstream tasks of classification, detection, and instance segmentation in most SSL previous works, we have extended the evaluation of our approach to include monocular depth estimation. To achieve this, we adopt Monodepth2 \cite{monodepth2} as our reference and substitute its feature encoder with our pretrained ResNet-50, which remains frozen during training. We maintain identical hyperparameters to those used in Monodepth2 and conduct our evaluation on the KITTI 2015 dataset \cite{Geiger2012CVPR}. The results are presented in Table \ref{tab:depth_estimation}. Notably, whereas Monodepth2 trains all model components, we exclusively train the depth decoder and the pose network while keeping the feature encoder fixed. Our mean filling masking strategy produces results on par with the original Monodepth2, and all our settings outperform baseline methods (MoCov2, MoCov2+MSCN, MoCov2+ADIOS). Furthermore, our approach's learned features demonstrates the capacity to generalize to tasks beyond the scope of traditional classification, detection, and instance segmentation.

\subsection{Computational Cost}\label{time_cost}
We compare the computational cost of ADIOS~\cite{shi2022adversarial}, MSCN~\cite{jing2022masked}, and our three masking strategies (i.e., high-pass filtering, strong blurring, and mean filling) using MoCov2 as the SSL framework. Training time (in minutes) per epoch in ImageNet-100 for each method is measured. Serving as the base SSL framework of all methods, MoCov2 takes 5.5 minutes to train one epoch. In order to alleviate the parasitic edges caused by masking operation in ConvNets, MSCN~\cite{jing2022masked} adopts a high-pass filter and applies random masking (including channel-wise and focal masking) on input images, which in results takes 7 minutes per epoch. Instead of randomly masking, ADIOS~\cite{shi2022adversarial} proposes an UNet-based occlusion module to adversarially learn along with the feature encoder to determine the regions to be masked, which is called masking slot. The memory and computational cost will increase linearly as the number of masking slots increases. 10 minutes are needed to train one epoch with 6 masking slots in ADIOS. In order to determine where and how to mask in an easier way, our three masking strategies consist of saliency computation and different image processing. In saliency computation, two forward passes through the localization network are needed to produce saliency maps for positive and (hard) negative samples. Compared to MSCN, although it takes 2 minutes longer per epoch due to the saliency constraint in our high-pass filtering strategy, we achieve better performance on various downstream tasks. While in mean filling and strong blurring strategies, mean value and strong blurred patches are filled in the masked regions to make those edges caused by masking less visible, in total each epoch takes 7.5 and 10.5 minutes respectively for their training. The strong blurring strategy spends more time than other strategies, in which the bottleneck is attributed to the GPU I/O. Since our saliency masking procedure is done on GPU, for our strong blurring strategy, we need to move both the standard augmented images and strong blurred images onto the GPU. The data transfer time will be twice that of our other two strategies  (i.e., high-pass filtering strategy only moves images onto GPU after high-pass filtering, while mean filling strategy only moves the images onto GPU after standard data augmentation). We will keep improving the overall GPU I/O procedure for our proposed strategies.
Furthermore, we test the accuracy of our high-pass filtering method against MSCN on ImageNet-100, with matching pre-training times. MSCN achieves its highest accuracy 70.28\% in 197 epochs, while around the same time our method based on high-pass filtering masking strategy reaches 131 epochs but results to have 71.66\% accuracy, which is already 1.4\% higher than MSCN.
To sum up, our high-pass filtering strategy strikes a better balance between efficiency and efficacy than MSCN and ADIOS.



%% file: main.bbl
\begin{thebibliography}{10}\itemsep=-1pt

\bibitem{bao2021beit}
Hangbo Bao, Li Dong, and Furu Wei.
\newblock Beit: Bert pre-training of image transformers.
\newblock In {\em ICLR}, 2022.

\bibitem{caron2018deep}
Mathilde Caron, Piotr Bojanowski, Armand Joulin, and Matthijs Douze.
\newblock Deep clustering for unsupervised learning of visual features.
\newblock In {\em ECCV}, 2018.

\bibitem{caron2020unsupervised}
Mathilde Caron, Ishan Misra, Julien Mairal, Priya Goyal, Piotr Bojanowski, and Armand Joulin.
\newblock Unsupervised learning of visual features by contrasting cluster assignments.
\newblock In {\em NeurIPS}, 2020.

\bibitem{chen2020simple}
Ting Chen, Simon Kornblith, Mohammad Norouzi, and Geoffrey Hinton.
\newblock A simple framework for contrastive learning of visual representations.
\newblock In {\em ICML}, 2020.

\bibitem{chen2020improved}
Xinlei Chen, Haoqi Fan, Ross Girshick, and Kaiming He.
\newblock Improved baselines with momentum contrastive learning.
\newblock {\em arXiv preprint arXiv:2003.04297}, 2020.

\bibitem{chen2021exploring}
Xinlei Chen and Kaiming He.
\newblock Exploring simple siamese representation learning.
\newblock In {\em CVPR}, 2021.

\bibitem{devlin2018bert}
Jacob Devlin, Ming-Wei Chang, Kenton Lee, and Kristina Toutanova.
\newblock Bert: Pre-training of deep bidirectional transformers for language understanding.
\newblock In {\em Proceedings of Annual Conference of the North American Chapter of the Association for Computational Linguistics: Human Language Technologies (NAACL-HLT)}, 2019.

\bibitem{dosovitskiy2020image}
Alexey Dosovitskiy, Lucas Beyer, Alexander Kolesnikov, Dirk Weissenborn, Xiaohua Zhai, Thomas Unterthiner, Mostafa Dehghani, Matthias Minderer, Georg Heigold, Sylvain Gelly, et~al.
\newblock An image is worth 16x16 words: Transformers for image recognition at scale.
\newblock {\em ArXiv:2010.11929}, 2020.

\bibitem{everingham2010pascal}
Mark Everingham, Luc Van~Gool, Christopher~KI Williams, John Winn, and Andrew Zisserman.
\newblock The pascal visual object classes (voc) challenge.
\newblock {\em IJCV}, 2010.

\bibitem{fei2004learning}
Li Fei-Fei, Rob Fergus, and Pietro Perona.
\newblock Learning generative visual models from few training examples: An incremental bayesian approach tested on 101 object categories.
\newblock In {\em IEEE Conference on Computer Vision and Pattern Recognition (CVPR) Workshops}, 2004.

\bibitem{feng2019self}
Zeyu Feng, Chang Xu, and Dacheng Tao.
\newblock Self-supervised representation learning by rotation feature decoupling.
\newblock In {\em CVPR}, 2019.

\bibitem{ge2021robust}
Songwei Ge, Shlok Mishra, Chun-Liang Li, Haohan Wang, and David Jacobs.
\newblock Robust contrastive learning using negative samples with diminished semantics.
\newblock In {\em NeurIPS}, 2021.

\bibitem{Geiger2012CVPR}
Andreas Geiger, Philip Lenz, and Raquel Urtasun.
\newblock Are we ready for autonomous driving? the kitti vision benchmark suite.
\newblock In {\em CVPR}, 2012.

\bibitem{gidaris2018unsupervised}
Spyros Gidaris, Praveer Singh, and Nikos Komodakis.
\newblock Unsupervised representation learning by predicting image rotations.
\newblock In {\em ICLR}, 2018.

\bibitem{monodepth2}
Cl{\'{e}}ment Godard, Oisin {Mac Aodha}, Michael Firman, and Gabriel~J. Brostow.
\newblock Digging into self-supervised monocular depth prediction.
\newblock In {\em ICCV}, October 2019.

\bibitem{grill2020bootstrap}
Jean-Bastien Grill, Florian Strub, Florent Altch{\'e}, Corentin Tallec, Pierre Richemond, Elena Buchatskaya, Carl Doersch, Bernardo Avila~Pires, Zhaohan Guo, Mohammad Gheshlaghi~Azar, et~al.
\newblock Bootstrap your own latent-a new approach to self-supervised learning.
\newblock In {\em NeurIPS}, 2020.

\bibitem{he2022masked}
Kaiming He, Xinlei Chen, Saining Xie, Yanghao Li, Piotr Doll{\'a}r, and Ross Girshick.
\newblock Masked autoencoders are scalable vision learners.
\newblock In {\em CVPR}, 2022.

\bibitem{he2020momentum}
Kaiming He, Haoqi Fan, Yuxin Wu, Saining Xie, and Ross Girshick.
\newblock Momentum contrast for unsupervised visual representation learning.
\newblock In {\em CVPR}, 2020.

\bibitem{he2017mask}
Kaiming He, Georgia Gkioxari, Piotr Doll{\'a}r, and Ross Girshick.
\newblock Mask r-cnn.
\newblock In {\em Proceedings of the IEEE international conference on computer vision}, pages 2961--2969, 2017.

\bibitem{jing2022masked}
Li Jing, Jiachen Zhu, and Yann LeCun.
\newblock Masked siamese convnets.
\newblock {\em ArXiv:2206.07700}, 2022.

\bibitem{Le2015TinyIV}
Ya Le and Xuan~S. Yang.
\newblock Tiny imagenet visual recognition challenge, 2015.

\bibitem{li2022semmae}
Gang Li, Heliang Zheng, Daqing Liu, Chaoyue Wang, Bing Su, and Changwen Zheng.
\newblock Semmae: Semantic-guided masking for learning masked autoencoders.
\newblock {\em Advances in Neural Information Processing Systems}, 2022.

\bibitem{li2021mst}
Zhaowen Li, Zhiyang Chen, Fan Yang, Wei Li, Yousong Zhu, Chaoyang Zhao, Rui Deng, Liwei Wu, Rui Zhao, Ming Tang, et~al.
\newblock Mst: Masked self-supervised transformer for visual representation.
\newblock {\em Advances in Neural Information Processing Systems}, 2021.

\bibitem{lin2014microsoft}
Tsung-Yi Lin, Michael Maire, Serge Belongie, James Hays, Pietro Perona, Deva Ramanan, Piotr Doll{\'a}r, and C~Lawrence Zitnick.
\newblock Microsoft coco: Common objects in context.
\newblock In {\em Computer Vision--ECCV 2014: 13th European Conference, Zurich, Switzerland, September 6-12, 2014, Proceedings, Part V 13}, pages 740--755. Springer, 2014.

\bibitem{nilsback2008automated}
Maria-Elena Nilsback and Andrew Zisserman.
\newblock Automated flower classification over a large number of classes.
\newblock In {\em Sixth Indian Conference on Computer Vision, Graphics \& Image Processing (ICVGIP)}, 2008.

\bibitem{noroozi2016unsupervised}
Mehdi Noroozi and Paolo Favaro.
\newblock Unsupervised learning of visual representations by solving jigsaw puzzles.
\newblock In {\em ECCV}, 2016.

\bibitem{oord2018representation}
Aaron van~den Oord, Yazhe Li, and Oriol Vinyals.
\newblock Representation learning with contrastive predictive coding.
\newblock In {\em NeurIPS}, 2019.

\bibitem{pathak2016context}
Deepak Pathak, Philipp Krahenbuhl, Jeff Donahue, Trevor Darrell, and Alexei~A Efros.
\newblock Context encoders: Feature learning by inpainting.
\newblock In {\em CVPR}, 2016.

\bibitem{peng2022crafting}
Xiangyu Peng, Kai Wang, Zheng Zhu, Mang Wang, and Yang You.
\newblock Crafting better contrastive views for siamese representation learning.
\newblock In {\em CVPR}, 2022.

\bibitem{radford2018improving}
Alec Radford, Karthik Narasimhan, Tim Salimans, Ilya Sutskever, et~al.
\newblock Improving language understanding by generative pre-training.
\newblock {\em OpenAI}, 2018.

\bibitem{ren2015faster}
Shaoqing Ren, Kaiming He, Ross Girshick, and Jian Sun.
\newblock Faster r-cnn: Towards real-time object detection with region proposal networks.
\newblock {\em Advances in neural information processing systems}, 28, 2015.

\bibitem{ruder2016overview}
Sebastian Ruder.
\newblock An overview of gradient descent optimization algorithms.
\newblock {\em arXiv preprint arXiv:1609.04747}, 2016.

\bibitem{ILSVRC15}
Olga Russakovsky, Jia Deng, Hao Su, Jonathan Krause, Sanjeev Satheesh, Sean Ma, Zhiheng Huang, Andrej Karpathy, Aditya Khosla, Michael Bernstein, Alexander~C. Berg, and Li Fei-Fei.
\newblock {ImageNet Large Scale Visual Recognition Challenge}.
\newblock {\em IJCV}, 2015.

\bibitem{shi2022adversarial}
Yuge Shi, N Siddharth, Philip Torr, and Adam~R Kosiorek.
\newblock Adversarial masking for self-supervised learning.
\newblock In {\em ICML}, 2022.

\bibitem{wang2022importance}
Xiao Wang, Haoqi Fan, Yuandong Tian, Daisuke Kihara, and Xinlei Chen.
\newblock On the importance of asymmetry for siamese representation learning.
\newblock In {\em CVPR}, 2022.

\bibitem{wei2017selective}
Xiu-Shen Wei, Jian-Hao Luo, Jianxin Wu, and Zhi-Hua Zhou.
\newblock Selective convolutional descriptor aggregation for fine-grained image retrieval.
\newblock {\em IEEE TIP}, 2017.

\bibitem{xie2022simmim}
Zhenda Xie, Zheng Zhang, Yue Cao, Yutong Lin, Jianmin Bao, Zhuliang Yao, Qi Dai, and Han Hu.
\newblock Simmim: A simple framework for masked image modeling.
\newblock In {\em CVPR}, 2022.

\bibitem{you2017large}
Yang You, Igor Gitman, and Boris Ginsburg.
\newblock Large batch training of convolutional networks.
\newblock {\em arXiv preprint arXiv:1708.03888}, 2017.

\bibitem{zbontar2021barlow}
Jure Zbontar, Li Jing, Ishan Misra, Yann LeCun, and St{\'e}phane Deny.
\newblock Barlow twins: Self-supervised learning via redundancy reduction.
\newblock In {\em ICML}, 2021.

\bibitem{zhang2016colorful}
Richard Zhang, Phillip Isola, and Alexei~A Efros.
\newblock Colorful image colorization.
\newblock In {\em ECCV}, 2016.

\bibitem{zhou2021ibot}
Jinghao Zhou, Chen Wei, Huiyu Wang, Wei Shen, Cihang Xie, Alan Yuille, and Tao Kong.
\newblock ibot: Image bert pre-training with online tokenizer.
\newblock In {\em ICLR}, 2022.

\end{thebibliography}
